\documentclass[conference]{IEEEtran}
\IEEEoverridecommandlockouts
\usepackage{cite}
\usepackage{amsmath,amssymb,amsfonts}
\usepackage{algorithmic}
\usepackage{graphicx}
\usepackage{textcomp}
\usepackage{xcolor}

\usepackage{CJKutf8}

\usepackage{booktabs}
\usepackage{multirow}
\def\BibTeX{{\rm B\kern-.05em{\sc i\kern-.025em b}\kern-.08em
    T\kern-.1667em\lower.7ex\hbox{E}\kern-.125emX}}
\begin{document}

\title{Teochew-Wild: The First In-the-wild Teochew Dataset with Orthographic Annotations}

\author{Linrong Pan$^1$ \\{\tt\small pain1kuan1@gmail.com}
\and Chenglong Jiang$^{1}$ \\{\tt\small csjiangcl\_gx@mail.scut.edu.cn}
\and Gaoze Hou$^2$ \\{\tt\small 314974247@qq.com}
\and Ying Gao$^{1*}$\thanks{*~Corresponding Author} \\{\tt\small gaoying@scut.edu.cn}
% \and $^1$School of Computer Science and Engineering, South China University of Technology, China\\
\\
\and \small $^1$School of Computer Science and Engineering, South China University of Technology, Guangzhou, China
\and \small ~~~~~~~~~~~~~~~~~~~~~~~~~~~~~~~~~~~~~~~~~~~~~~$^2$Guangzhou No.6 Middle School, Guangzhou, China
% \\
% School of Computer Science and Engineering, South China University of Technology, China
}

\maketitle

\begin{abstract}

This paper reports the construction of the Teochew-Wild, a speech corpus of the Teochew dialect. The corpus includes 18.9 hours of in-the-wild Teochew speech data from multiple speakers, covering both formal and colloquial expressions, with precise orthographic and pinyin annotations. Additionally, we provide supplementary text processing tools and resources to propel research and applications in speech tasks for this low-resource language, such as automatic speech recognition (ASR) and text-to-speech (TTS). To the best of our knowledge, this is the first publicly available Teochew dataset with accurate orthographic annotations. We conduct experiments on the corpus, and the results validate its effectiveness in ASR and TTS tasks.

\end{abstract}


\section{Introduction}
\label{sec:intro}
With the rapid advancements in deep learning and the availability of large-scale datasets \cite{zhang2022wenetspeech,emilia,ardila2020common}, state-of-the-art Automatic Speech Recognition (ASR) systems now deliver high-precision transcription for major languages, including English, Mandarin Chinese, Japanese, and others \cite{radford2023robust,bai2024seed,jeffries2024moonshinespeechrecognitionlive}. Simultaneously, cutting-edge Text-to-Speech (TTS) systems \cite{wang2024maskgct,wang2023neural,du2024cosyvoice,chen2024f5} have achieved the capability to synthesize highly natural and expressive speech for these languages.

In contrast, many languages worldwide remain severely under-resourced, with Teochew being a prominent example. As a branch of the Southern Min (Hokkien) language family, Teochew is predominantly spoken in eastern Guangdong Province, China, and various regions of Southeast Asia. Rough estimates indicate that the global Teochew-speaking population is approximately 30 million. Despite this relatively large speaker base, publicly available resources for Teochew are exceedingly scarce. Notably, Teochew is absent even from large-scale speech datasets \cite{black2019cmu,pratap2024scaling} that include hundreds or even thousands of languages.

We posit that one of the primary reasons for this scarcity is the widespread misconception among researchers that Teochew and Hokkien are the same language. Consequently, prior studies have predominantly focused on Hokkien, often neglecting the dedicated collection and analysis of Teochew. In reality, Teochew and Hokkien exhibit substantial linguistic differences and are largely mutually unintelligible. Previous studies \cite{magistry2024experiments} attempted to augment Teochew speech synthesis models by combining context-free Teochew data (comprising only isolated characters or words) with larger-scale Taiwanese Hokkien data. Nevertheless, experimental results revealed that incorporating Taiwanese Hokkien data did not significantly improve the performance of Teochew speech synthesis.

In recent years, numerous studies have focused on speech recognition and synthesis for low-resource languages, achieving notable success \cite{baevski2020wav2vec,hsu2021hubert,chen2022wavlm,hsu2023low,zhao2022improving,xu2020lrspeech}. However, the solutions proposed in these studies still require at least 10 hours of annotated speech data. To the best of our knowledge, apart from some small-scale commercial datasets (whose annotation quality is unsatisfactory), there is currently no publicly available Teochew dataset with accurate orthographic annotations in the academic domain.

To address this research gap, we construct the first Teochew speech corpus with precise orthographic annotations, named Teochew-Wild. Additionally, we leverage the collected orthographic and lexical data to enhance and refine the Teochew writing system and text front-end. The corpus data is entirely sourced from the Internet, comprising 20 speakers and 12,500 utterances. Both Chinese characters and Teochew pinyin are used for annotation, enabling the corpus to be applied to various speech-related tasks such as ASR and TTS. The main contributions of this paper are as follows:

\begin{figure*}[t]
\setlength{\abovecaptionskip}{0pt}
    \centering
    \includegraphics[scale=0.85]{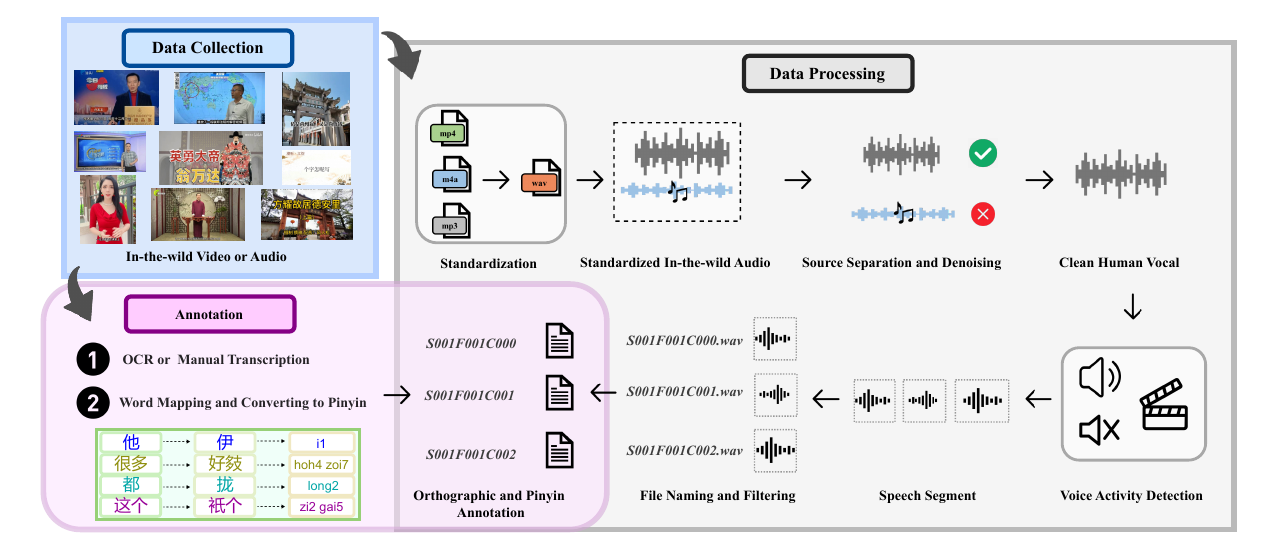}
    \caption{The Overview of Our Dataset Construction Workflow}
    \label{fig:pipeline}
\end{figure*}

\vspace{1mm}
\begin{itemize}

  \item We introduce the first Teochew corpus annotated with both standard Chinese characters and Teochew pinyin. This corpus includes formal written expressions as well as authentic colloquial speech, exhibiting a high degree of diversity and naturalness.
  \item We have developed text processing tools and resources for Teochew, which facilitate polyphonic character disambiguation, Mandarin-to-Teochew mapping, and Grapheme-to-Phoneme (G2P) conversion. These tools address the lack of text frontend resources for Teochew speech synthesis.
  \item We have refined and enriched the Teochew writing system, ensuring that every pronunciation is linked to a corresponding Chinese character. This enhancement supports neural models in achieving better performance on this language.
  \item Our experiments demonstrate the effectiveness of the Teochew-Wild in speech tasks, which will facilitate the construction of large-scale datasets in the future.

\end{itemize}
% \vspace{-1mm}
The dataset and related resources are available at our repository\footnote{https://huggingface.co/datasets/panlr/teochew\_wild}.

\section{Background}

Teochew (also known as Teo-Swa, Chaozhou dialect) is a unique subset of the Southern Min dialect group. Under the influence of various external factors, Teochew has undergone independent linguistic evolution, resulting in significant differences from Southern Min spoken in Fujian or Taiwan. Due to the widespread promotion of official languages in China and Southeast Asian countries, the younger generation's proficiency in Teochew is gradually declining, both in local communities and among the diaspora, pushing the language toward endangerment. At present, most Teochew speakers only possess basic oral communication skills, while remaining unfamiliar with many idioms, colloquialisms, and formal vocabulary. This deterioration of oral proficiency and limited literacy underscores the urgency for systematic research and preservation efforts for the Teochew dialect.

\subsection{The Writing System of Teochew}

Due to the absence of standardization and promotion by authoritative organizations, Teochew lacks a widely accepted standardized writing system. As a Chinese dialect, Teochew can be partially represented using Chinese characters. However, because its high-frequency vocabulary differs significantly from Mandarin and many colloquial words lack corresponding Chinese characters, even native speakers are not accustomed to using a formal orthographic system.

In contrast, the phonetic system of Teochew is relatively more developed. In 1960, the Guangdong provincial education authorities introduced Teochew Pinyin (Pengim), which consists of 17 initials and 59 finals. This phonological system serves as a foundation for our dataset construction.

% \vspace{-3mm}
\subsection{Tone Sandhi in Teochew}
Teochew features eight fundamental tones (in comparison, Mandarin has four, and Fujian's Southern Min has seven). When continuous reading, the tones change depending on their context, a phenomenon known as tone sandhi. Due to the larger number of fundamental tones, tone sandhi in Teochew is more complex than in Mandarin or Southern Min. Although certain regular patterns exist, these rules are insufficient to account for all sandhi variations. Consequently, Teochew linguistic resources generally adopt annotations based on the original tones.

\subsection{Polyphone in Teochew}
Similar to other Southern Min dialects, Teochew exhibits a phenomenon of literary vs. vernacular pronunciations. That is, some Chinese characters are pronounced with literary pronunciation in contexts such as reading formal texts, poetry, or persons' names, while the vernacular pronunciation is used in everyday communication or ``earthier'' language. 
Due to the presence of literary and vernacular pronunciations, the polyphonic characters in Teochew are more complex than in Mandarin. For example, the character ``\begin{CJK}{UTF8}{gbsn}相\end{CJK}'' has two pronunciations in Mandarin, but in Teochew, these two pronunciations correspond to both literary and vernacular pronunciation, resulting in four distinct pronunciations.

\section{Teochew-Wild Corpus}
In this section, we will describe the process of constructing the Teochew-Wild Corpus (as shown in Fig.~\ref{fig:pipeline}), as well as our data annotation methodology.

\subsection{Data Collection}
As a predominantly spoken language, Teochew contains many words and phrases that are not documented in written materials. To collect more diverse, rich, and authentic spoken data reflecting everyday usage, we chose to gather in-the-wild data from the internet rather than recording in a studio. The richness of in-the-wild data contributes to improving the Teochew writing system and text frontend, thereby better supporting the application of speech technologies in this dialect. 

We search for Teochew multimedia content online, prioritizing audio or video with minimal noise and sufficient duration. Ultimately, we collect over 200 hours of raw speech data.

\subsection{Data Preprocessing}
From the raw data, we select 20 speakers with clear pronunciation and sufficient speech duration (11 males and 9 females) and extract approximately 20 hours of data for annotation. The remaining data is utilized as unsupervised data in future research.
Next, we adopt and modify the data preprocessing pipeline proposed by \cite{emilia} to automate the processing of these raw data. The improved pipeline, illustrated in Figure 1, comprises the following five steps:

a) Speaker Preprocessing. We prioritize data containing a single speaker. If the data includes non-target speakers or overlapping speech, we use editing tools to remove these segments. This ensures that speaker diarization is not required in subsequent processing.

b) Standardization. In-the-wild speech data vary in encoding formats and sampling rates. Following the method in \cite{emilia}, we standardize all samples by converting them to WAV format, setting them to mono-channel, and resampling at 22050 Hz.

c) Source Separation and Denoising. In-the-wild speech data often include background music or other noise, which can negatively impact speech synthesis tasks. We apply the Ultimate Vocal Remover (UVR)\footnote{https://github.com/Anjok07/ultimatevocalremovergui} model for source separation to effectively extract vocal tracks from background music. Furthermore, for audio that remains noisy after source separation, we employ the Resemble-Enhance\footnote{https://github.com/resemble-ai/resemble-enhance} model for secondary denoising.

d) Voice Activity Detection (VAD). We use the Silero-VAD \footnote{https://github.com/snakers4/silero-vad} model to split speech data into smaller VAD segments. Consecutive VAD segments are merged into utterances of appropriate lengths, ensuring each utterance is no shorter than 1 second and no longer than 20 seconds. Based on the VAD results, the audio is segmented into multiple clips and named using the format ``$S{_{sid}}F{_{fid}}C{_{cid}}.wav$'', where S represents the Speaker, F represents the File, and C represents the Clip.

e) Filtering. We select a pure noise segment to calculate the signal-to-noise ratio (SNR) of each audio clip. Clips with SNR values below the predefined threshold are automatically filtered out to ensure data quality.

\begin{table}[t]
    \centering
    \setlength{\abovecaptionskip}{0pt}
    \caption{Several examples to illustrate the mapping relationship between Teochew and Mandarin vocabulary.}
    \begin{tabular}{c c c}
     \hline
        Meaning & Mandarin Word  & Teochew  Word \\ \hline
        they  & \begin{CJK}{UTF8}{gbsn}他们\end{CJK}  & \begin{CJK}{UTF8}{gbsn}伊人\end{CJK}  \\  
        where & \begin{CJK}{UTF8}{gbsn}哪里\end{CJK} &  \begin{CJK}{UTF8}{gbsn}底块\end{CJK}  \\ 
        small	& \begin{CJK}{UTF8}{gbsn}小\end{CJK} 	&  \begin{CJK}{UTF8}{gbsn}细\end{CJK} \\  
        eat	&  \begin{CJK}{UTF8}{gbsn}吃\end{CJK}	&  \begin{CJK}{UTF8}{gbsn}食\end{CJK}  \\  
        walk	&   \begin{CJK}{UTF8}{gbsn}走\end{CJK} 	&  \begin{CJK}{UTF8}{gbsn}行\end{CJK}   \\ 
        same   &  \begin{CJK}{UTF8}{gbsn}同样\end{CJK}  & \begin{CJK}{UTF8}{gbsn}平样\end{CJK}  \\ \hline

    \end{tabular}
    
    \label{tab:example}
\end{table}
\subsection{Text Frontend}
The primary function of the text frontend is to convert input text into linguistic features, including pronunciation and prosody, facilitating speech synthesis. Other functionalities can leverage existing Mandarin text frontend frameworks, so we focus on developing the following three functional modules:
a) Grapheme-to-phoneme conversion;
b) Disambiguation of polyphonic characters;
c) Mapping between Teochew and Mandarin words.

\subsubsection{Grapheme-to-phoneme}
Due to the lack of support from authoritative institutions, the Grapheme-to-Phoneme system for Teochew has long remained underdeveloped. To address this issue, we first crawled all Chinese characters and their corresponding pronunciations from a community-created online Teochew dictionary\footnote{http://www.czyzd.com/}. Additionally, we consulted relevant resources\footnote{https://learnteochew.com/pages/pronunciation.html} to construct a lexicon that supports the decomposition of pinyin into phonemes and facilitates the conversion of Teochew phonemes into the International Phonetic Alphabet (IPA).

\subsubsection{Polyphone Disambiguation}
We collected a total of 9,143 Chinese characters along with their corresponding pronunciations, including 2,256 polyphonic characters, 364 of which have more than three pronunciations. This complexity of polyphonic characters presents significant challenges for converting Chinese characters into pinyin. Through observation, we found that many polyphonic characters have only one or two commonly used pronunciations, while others are infrequent. Even those with multiple high-frequency pronunciations usually have a fixed pronunciation in specific word groups.

Based on this pattern, we construct a lexicon containing over 10,000 word groups. Each word group includes at least one polyphonic character and maps to its possible pinyin combinations. This rule-based approach is a simple yet practical solution for polyphone disambiguation, significantly reducing the labor required for subsequent annotation tasks.

\subsubsection{Dialect Words Mapping}
Teochew exhibits significant differences from Standard Mandarin in word usage and includes a substantial number of region-specific vocabulary items, resulting in considerable textual divergence between the two languages (see Table \ref{tab:example}). To address this issue, we develop a mapping dictionary containing over 1,500 entries, which converts text vocabulary into more native Teochew expressions. This mapping dictionary not only facilitates annotation but also enables speech generation models to produce audio that more closely reflects spoken Teochew.

\subsection{Annotation}
Skilled annotators proficient in Teochew orthography and phonetic transcription are extremely scarce. To balance annotation quality and efficiency, we design a two-stage annotation process consisting of coarse annotation and fine-grained annotation. In the coarse annotation stage, the speech is transcribed into Mandarin text with semantically similar meanings. In the fine-grained annotation stage, the Mandarin text is further mapped to Teochew orthography and phonetic transcription.

During the first annotation phase, we engage four native Teochew speakers to organize the data. For data containing subtitles, Optical Character Recognition (OCR) technology is utilized to extract text, which annotators then review to correct errors and remove irrelevant information. For data without subtitles, annotators directly transcribe the speech. To standardize the annotation process, we provide the Teochew-Mandarin vocabulary mapping dictionary and instruct annotators to use the provided vocabulary whenever possible.

The second annotation phase is carried out by two experts proficient in Teochew orthography and phonetic transcription.
Leveraging the text-processing tools we developed, they perform precise annotations for each pronunciation, strictly adhering to the following annotation rules:

\begin{itemize}
  \item  For simple English abbreviations (e.g., CCTV, APP) in the audio, retain them and convert them into phonetically equivalent pinyin;
  \item Convert Arabic numerals into Chinese characters based on their pronunciation;
  \item Retain the original tone for tonal annotations unless special tone sandhi is used by the speaker;
  \item Adjust phonetic transcriptions based on the speaker's regional accent.
  \end{itemize}
\subsection{Supplementation of the Writing System}
During the fine-grained annotation stage, we frequently encounter pronunciations that do not correspond to existing Chinese characters. To resolve this, we select rarely used characters from ancient Chinese texts and assign them new pronunciations and meanings, thereby supplementing and refining the writing system.

\section{Corpus Details}
We randomly divide the Teochew-Wild dataset into training, validation, and testing sets, consisting of 11,100, 700, and 700 samples, respectively. Next, we analyze the corpus in terms of speech quality and data statistics.

\begin{table}[t]
\setlength{\abovecaptionskip}{0pt}
\centering
\caption{Quality comparison between Teochew-Wild and six existing TTS datasets. The scores for Aishell-3 and LibriTTS are derived from \cite{yu2024autoprep}, while the scores for VCTK, GigaSpeech, WenetSpeech4TTS, and Emilia are sourced from \cite{emilia}.}
\label{tab:dnsmos_results}
\begin{tabular}{ccc}
\toprule
 
\textbf{Dataset} & \textbf{DNSMOS} & \textbf{Data Source} \\  
\midrule
VCTK~\cite{yamagishi2019cstr} & 3.20 ± 0.18 & Studio Recording \\ 
Aishell-3~\cite{aishell3} & 3.15 ± 0.17 & Studio Recording \\ 
LibriTTS~\cite{zen2019libritts} & 3.25 ± 0.19 & Audiobook \\  
GigaSpeech~\cite{chen2021gigaspeech} & 2.52 ± 0.54 & In-the-wild \\
WenetSpeech4TTS~\cite{ma2024wenetspeech4tts} & 3.18 ± 0.22 & In-the-wild \\
Emilia~\cite{emilia} & 3.26 ± 0.14 & In-the-wild \\ 
\midrule
Teochew-Wild & 3.12 ± 0.32 & In-the-wild \\
\bottomrule
\end{tabular}
\end{table}

\begin{table}[t]
\setlength{\abovecaptionskip}{0pt}
\caption{The duration and character count statistics of the Teochew-Wild corpus.}
\begin{tabular}{ccccc}

\hline
metric               & min  & max   & avg ± std      & total  \\ \hline
duration(seconds)    & 1.17 & 18.82 & 5.46 ± 2.55  & 68053  \\
number of characters & 3    & 75    & 21.73 ± 9.52 & 271625 \\ \hline
\end{tabular}
\label{tab:statistic}
\end{table}

\subsection{Speech Quality}
To evaluate speech quality, we compare Teochew-Wild with existing datasets using DNSMOS P.835 OVRL scores \cite{reddy2022dnsmos}. This non-intrusive speech quality metric reflects the overall quality of speech data and shows a high correlation with human ratings. Table \ref{tab:dnsmos_results} presents the speech quality comparison between Teochew-Wild and several existing datasets.
Although the Teochew-Wild dataset is derived from raw field-collected speech data, its DNSMOS P.835 OVRL score of 3.12 indicates that the speech quality is comparable to that of other datasets sourced from in-the-wild data, studio recordings, or audiobooks. This ensures the dataset's effectiveness for speech synthesis applications.

\subsection{Data Statistics}
Teochew-Wild comprises 18.9 hours of Teochew speech data from 20 speakers, totaling 12,500 samples. As summarized in Table \ref{tab:statistic}, each audio sample has an average duration of 5.45 seconds and contains approximately 21.7 characters. The sample durations range from 1.17 to 18.82 seconds, with the majority being short clips between 2 and 9 seconds.

Statistical analysis reveals that the Teochew-Wild dataset includes a total of 271,625 Chinese characters, consisting of 3,853 unique characters. It covers 882 initial and final combinations, with a coverage rate of 94.8\%. The results indicate that a small subset of characters occurs far more frequently than others. Specifically, about 12.13\% of the characters in the Teochew dataset are made up of the 10 most common characters. At the same time, there are over 1,800 characters that appear fewer than 10 times, reflecting the textual diversity within the dataset.

Our Teochew-Wild dataset spans multiple domains, including topics such as local news, current affairs commentary, storytelling (Pingshu), history, local cultural introductions, international relations, storybooks, poetry, and more. It encompasses both formal written language and a large number of words commonly used in daily life, demonstrating significant diversity.

\section{EXPERIMENTS}
In this section, we conduct TTS and ASR experiments on our Teochew-Wild dataset to validate its effectiveness. Given that state-of-the-art TTS and ASR models typically require thousands of hours of training data to converge, we selected models that are suitable for smaller datasets for verification. Specifically, in the TTS experiment, we used the autoregressive (AR) model Tacotron2 \cite{shen2018natural} and the non-autoregressive (NAR) model FastSpeech2 \cite{ren2020fastspeech} to predict mel-spectrograms, with the HiFi-GAN \cite{kong2020hifi} vocoder used to convert them into waveforms. In the ASR experiment, we trained the Fairseq S2T Transformer XS \cite{wang2020fairseq} with both character-based and pinyin-based annotations. Additionally, we fine-tuned the Whisper-medium pre-trained model to assess the applicability of our dataset to large models.

\subsection{Experiment Setup}
In the TTS experiment, both the Tacotron2 and FastSpeech2 models are trained for 30,000 steps on the Teochew-Wild training set, while HiFi-GAN is pre-trained on 200 hours of unlabeled data. In the ASR experiment, we downsample the audio to 16,000 Hz and use a toolkit to train the Fairseq S2T Transformer model and Whisper-medium until convergence. During testing, we select the checkpoint that performs best on the validation set. Other hyperparameters follow the default configuration as closely as possible.

\subsection{Evaluation Metric}
\subsubsection{TTS experiment}
Considering that Teochew is an unseen language for existing objective evaluation models, we conducted only subjective evaluations. We randomly selected 50 utterances from the test set and evaluated them using the Mean Opinion Score (MOS) on a 5-point scale. Twelve native Teochew speakers (aged 18 to 60) participated in the evaluation, with each tester listening to 20 audio samples. The testers were instructed to provide a comprehensive evaluation based on both the intelligibility and naturalness of the speech.

\subsubsection{ASR experiment}
We assess the transcription performance of the models on the test set. For models using character-based annotations, we employ Character Error Rate (CER) as the evaluation metric; for models using pinyin annotations, we use Word Error Rate (WER) as the evaluation metric.

\subsection{Experiment Results}
\subsubsection{TTS}
The results of the TTS experiments are shown in Table \ref{tab:MOS}. Despite the relatively small size of our dataset and its in-the-wild source, the MOS scores of the synthesized audio generated by Tacotron2 and FastSpeech2 reach 3.52 and 3.22, respectively, and are intelligible to the testers. According to the testers' feedback, compared to NAR models, AR models learn more contextual information, resulting in better performance in tone sandhi.

\subsubsection{ASR}
The experimental results of ASR are presented in Table \ref{tab:CER}. The results demonstrate the effectiveness of these models in Teochew dialect speech recognition tasks, regardless of whether the annotations are in Chinese characters or pinyin. Due to the large number of homophones in Chinese characters, the CER of Fairseq S2T Transformer XS is relatively high with Chinese character annotations. When pinyin annotations are used, the CER decreases significantly to 14.60\%(validation set) and 16.88\% (testing set). Notably, the Whisper model, after fine-tuning for 10 epochs on Teochew it had not previously encountered, achieved highly competitive results on both the validation and testing sets, with CERs of 9.61\% and 10.01\%, respectively. This shows that large models exhibit robust generalization and adaptability for low-resource languages.

\begin{table}[t]
\setlength{\abovecaptionskip}{0pt}
\centering
\caption{MOS with 95\% confidence intervals for the text-to-speech experiments.}

\begin{tabular}{ll}
\hline
\textbf{Method}    & \textbf{MOS}   \\ \hline
GT    & 4.33 ± 0.19      \\
GT Mel + HiFi-GAN   & 4.18 ± 0.20      \\
Tacotron 2  + HiFi-GAN &  3.52 ± 0.25    \\
FastSpeech 2  + HiFi-GAN   & 3.22 ± 0.24     \\ \bottomrule
\end{tabular}
\label{tab:MOS}
\end{table}

\begin{table}[t]
\centering
\setlength{\abovecaptionskip}{0pt}
\caption{The evaluation results (WER \& CER) of Fairseq S2T Transformer XS and Whisper-Medium on the validation and test sets. The Whisper model was fine-tuned using an existing pre-trained model.}
\begin{tabular}{llll}
\hline
\multirow{2}{*}{Model} & \multirow{2}{*}{Annotation Type} & \multicolumn{2}{c}{Subset} \\ \cline{3-4} 
                       &                                  & val          & test        \\ \hline
Fairseq s2t XS         & Chinese Character               & 35.34         & 39.36        \\
Fairseq s2t XS         & Teochew Pinyin                   & 14.60        & 16.88       \\
Whisper-Medium(ft)         & Chinese Character                & 9.61         & 10.01       \\ 
Whisper-Medium(ft)         & Teochew Pinyin                & 14.39         &  15.03       \\ 
\hline
\end{tabular}
\label{tab:CER}
\end{table}

\section{CONCLUSION}
In this work, we introduce the Teochew-Wild corpus with orthographic annotations and the necessary text front-end modules for speech synthesis, marking an initial step toward Teochew speech recognition and synthesis. Preliminary experiments validate the effectiveness of the dataset. Nonetheless, compared to mainstream datasets, the scale of our data remains relatively small, and there is still a gap before it can be applied to real-world scenarios. Therefore, in the future, we will continue to explore this dialect, leveraging speech recognition and synthesis methods in low-resource settings to continuously expand the relevant data and resources.

\section{ACKNOWLEDGEMENTS}
This work was supported in part by the Guangzhou Key Research and Development Program under Grant No.2024B01W0029.

\bibliographystyle{IEEEbib}
\bibliography{icme2025references}

\end{document}